\documentclass[10pt,twocolumn,letterpaper]{article}

\usepackage[pagenumbers]{cvpr}   
\definecolor{cvprblue}{rgb}{0.21,0.49,0.74}
\usepackage[pagebackref,breaklinks,colorlinks,allcolors=cvprblue]{hyperref}
\usepackage{graphicx}
\usepackage{booktabs}
\usepackage{amsmath}
\usepackage{array}
\usepackage{listings}
\usepackage{xcolor}
\def\paperID{*****} 
\def\confName{CVPR}
\def\confYear{2026}

\title{WADE: A Reasoning-Annotated Benchmark for Multi-Instance Floating-Waste Grounding with Compact Vision-Language Models}

\author{
\begin{tabular}{c}
Md. Asaduzzaman Shuvo,\;
Ahsan Farabi,\;
Md. Abdul Ahad Minhaz,\;
Mahedi Hasan\textsuperscript{\textdagger},\\
Israt Khandaker\textsuperscript{\textdagger},\;
Ibrahim Khalil Shanto,\;
Muhammad Nomani Kabir\textsuperscript{*}
\end{tabular}
\\
\normalsize
United International University
\\
\resizebox{0.95\textwidth}{!}{%
\textbf{Emails: }
\texttt{\{ashuvo221104,afarabi221266,mminhaz213072,mhasan221119,ikhandaker221263,ishanto213193\}@bscse.uiu.ac.bd}%
}\\
\small
\textsuperscript{*}\textit{Corresponding author:}
Muhammad Nomani Kabir,
\texttt{kabir@cse.uiu.ac.bd}\\
\small
\textdagger  contributed equally to this work.
}

\date{}

\begin{document}

\maketitle

\begin{abstract}
Floating waste in inland waterways threatens aquatic ecosystems and requires timely monitoring under cluttered, multi-object conditions. Existing aquatic-waste datasets provide limited geographic coverage, sparse multi-instance annotations, and little supervision beyond boxes and labels. Compact vision-language models (VLMs) therefore remain insufficiently evaluated for jointly localizing, classifying, counting, and explaining floating waste. We introduce \textbf{WADE}, a reasoning-annotated benchmark containing 2,167 images from rural Bangladesh, 13,608 bounding boxes, and ten waste categories. Each annotation is associated with class-level recognition rules covering visual cues, likely confusions, and discriminative features. We evaluate six VLMs under zero-shot, two-shot, reasoning-guided, and fine-tuned settings using detection, counting, and hallucination metrics. For resource-efficient adaptation, we jointly fine-tune Qwen3-VL-2B on boxes, labels, and reasoning chains using QLoRA. Fine-tuning increases recall from 0.0248 to 0.2339 and F1 from 0.0257 to 0.2163, while reducing image-level hallucination from 0.6836 to 0.0883. However, over three-quarters of instances remain undetected, establishing WADE as a challenging benchmark for dense floating-waste grounding with compact VLMs.
\end{abstract}

\section{Introduction}
\label{sec:introduction}

Floating waste in ponds, rivers, and canals threatens aquatic ecosystems and contributes to the movement of land-based pollution into marine environments \cite{jambeck2015plastic,meijer2021rivers}.
\begin{figure}[t]
    \centering
    \includegraphics[width=\columnwidth]{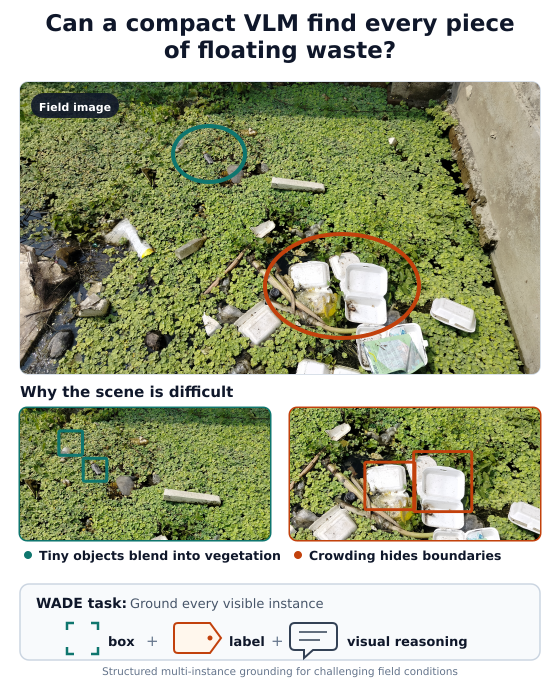}
    \caption{Overview of the WADE task. Floating-waste instances can be
    small, crowded, partially occluded, and visually similar to surrounding
    vegetation. Given a field image, a compact vision-language model must
    ground each visible instance and provide its bounding box, category
    label, and visual reasoning.}
    \label{fig:wade_teaser}
\end{figure}
Computer vision offers a scalable alternative to manual monitoring, and datasets such as TACO \cite{proenca2020taco} and FloW \cite{cheng2021flow} have supported litter and floating-waste detection. Nevertheless, real water-surface images remain challenging because waste objects are often small, visually similar to vegetation, partially submerged, overlapping, and unevenly distributed across the scene, as illustrated in Fig.~\ref{fig:wade_teaser}.
Vision-language models (VLMs) can combine recognition, spatial grounding, and textual explanation within a single generative interface \cite{liu2024llava,bai2025qwen3vl}. However, identifying one salient object is different from grounding every visible instance. In cluttered scenes, VLMs may miss small objects, generate inaccurate boxes, or predict categories unsupported by the image. Such object hallucination is a known limitation of generative vision-language systems \cite{rohrbach2018chair,li2023pope}. Visual reasoning annotations may help connect predictions with discriminative image evidence \cite{shao2024visualcot}, but their role in multi-instance floating-waste grounding remains underexplored.

We introduce \textbf{WADE}, a reasoning-annotated benchmark for multi-instance floating-waste grounding with compact VLMs. WADE contains 2,167 in-situ images from rural Bangladeshi waterways, 13,608 bounding boxes, and ten categories, with an average of 6.28 instances per image. Each instance is associated with a structured \emph{class-level} reasoning chain describing a primary visual cue, a confusable category, and a separating rule. We evaluate six VLMs under zero-shot, two-shot, reasoning-prompted, and fine-tuned settings. Joint QLoRA fine-tuning \cite{dettmers2023qlora} of Qwen3-VL-2B increases recall from 0.0248 to 0.2339 and class-aware F1 from 0.0257 to 0.2163, while reducing image-level hallucination from 0.6836 to 0.0883. However, most annotated instances remain undetected, showing that the task is still far from solved. Because boxes, labels, and reasoning are used jointly during fine-tuning, these improvements should not be attributed to reasoning alone. Our contributions are:

\begin{itemize}

\item \textbf{WADE dataset:} A real-world multi-instance
floating-waste benchmark containing 2,167 images, 13,608
bounding-box annotations, and ten categories collected from
rural waterways in Bangladesh.

\item \textbf{Reasoning annotations:} A structured class-level
reasoning schema that captures primary visual cues, likely
confusions, contrastive rules, and fallback evidence for each
waste category.

\item \textbf{Comprehensive VLM evaluation:} A systematic
evaluation of six VLMs under zero-shot, two-shot,
reasoning-guided, and QLoRA fine-tuned settings using grounding,
counting, and hallucination metrics.

\item \textbf{Empirical findings:} Evidence that joint in-domain
supervision improves Qwen3-VL-2B recall from 0.0248 to 0.2339
and reduces image-level hallucination from 0.6836 to 0.0883,
while showing that exhaustive grounding in cluttered waterways
remains unresolved.

\end{itemize}

\section*{Data and Code Availability}
The WADE dataset, annotations, split files, prompts, evaluation code, and fine-tuning implementation will be made publicly available upon acceptance.

\section{Related Work}
\label{sec:related_work}

Previous research has studied waste detection across several imaging platforms. A camera-based system for monitoring floating plastics in rivers demonstrated the feasibility of continuous automated observation while identifying reflections, organic material, and environmental variation as important challenges~\cite{vanlieshout2020automated}. At a substantially larger scale, MARIDA introduced multispectral Sentinel-2 imagery with pixel-level annotations for distinguishing marine debris from water, vegetation, foam, ships, and other surface features~\cite{kikaki2022marida}. Waste detection has also been investigated in highly cluttered industrial environments. The ZeroWaste dataset, for example, focuses on detecting and segmenting deformable, overlapping, and partially transparent waste objects~\cite{bashkirova2022zerowaste}. These studies primarily formulate waste analysis as detection or segmentation. In contrast, WADE considers multi-instance grounding in low-resolution waterway images, requiring compact vision-language models to enumerate visible waste objects and jointly produce their locations, labels, and supporting descriptions.

Language-guided detection connects visual localization with natural-language concepts \cite{kamath2021mdetr}. GLIP unifies object detection and phrase grounding through language-aware pre-training~\cite{li2022glip}, while Grounding DINO extends this direction to open-set localization using category names or referring expressions~\cite{liu2024groundingdino}. More recently, GLaMM integrates multimodal generation with pixel-level grounding, enabling textual responses that are linked to localized image regions~\cite{rasheed2024glamm}. These methods demonstrate strong general-purpose grounding capabilities, but environmental monitoring introduces additional difficulties: objects may be small, visually degraded, densely distributed, or easily confused with reflections and natural debris. WADE therefore evaluates whether compact generative vision-language models can produce a complete structured response containing multiple boxes, category labels, and short category-level rationales under such conditions.

Environmental datasets are increasingly incorporating language in addition to conventional visual annotations. Most closely related to our annotation perspective, VisText-Mosquito combines object detection, water-surface segmentation, and natural-language explanations for identifying mosquito breeding environments~\cite{islam2026vistextmosquito}. It shows the value of connecting environmental observations with interpretable textual information rather than providing spatial annotations alone. WADE explores this idea in a different setting: dense floating-waste scenes in rural waterways. Its annotations associate multiple object-level bounding boxes and waste labels with category-level discriminative rationales, supporting evaluation across detection, recognition, structured response generation, and explanation consistency. Thus, WADE complements existing environmental multimodal resources by targeting exhaustive multi-instance grounding with compact vision-language models.

\section{The WADE Dataset}
\label{sec:dataset}

WADE is a multi-instance grounding dataset for floating-waste recognition in
unconstrained waterways. It contains 2,167 field images and 13,608 annotated
instances from ten categories, with an average of 6.28 instances per image.
Each instance has a bounding box and category label and is associated with a
validated class-level reasoning chain.

\subsection{Collection and Taxonomy}
\label{sec:collection_taxonomy}

Images were captured using consumer smartphones across ponds, canals, and
rivers at 390 geographic locations in rural Bangladesh. The collection covers
winter and monsoon conditions together with morning, midday, and low-light
scenes. No waste was placed or rearranged during collection; all annotated
objects occurred naturally in the waterways.

The taxonomy comprises \textit{water hyacinth}, \textit{plastic bottle},
\textit{organic waste}, \textit{industrial waste}, \textit{polythene},
\textit{algal bloom}, \textit{solid waste}, \textit{fabric waste},
\textit{wood debris}, and \textit{foam waste}. Biological surface materials
are included because they frequently coexist with waste and form difficult
visual confounders. Table~\ref{tab:wade_stats} summarizes the dataset.

\begin{table}[t]
    \centering

    \scriptsize
    \setlength{\tabcolsep}{2.8pt}
    \begin{tabular}{lrr@{\hspace{5pt}}lrr}
        \toprule
        \textbf{Class} & \textbf{N} & \textbf{\%} &
        \textbf{Class} & \textbf{N} & \textbf{\%} \\
        \midrule
        Water hyacinth   & 2,137 & 15.7 & Algal bloom  & 1,110 & 8.2 \\
        Plastic bottle   & 1,933 & 14.2 & Solid waste  & 1,008 & 7.4 \\
        Organic waste    & 1,605 & 11.8 & Fabric waste &   976 & 7.2 \\
        Industrial waste & 1,563 & 11.5 & Wood debris  &   887 & 6.5 \\
        Polythene        & 1,525 & 11.2 & Foam waste   &   864 & 6.3 \\
        \midrule
        \multicolumn{3}{l}{Images: 2,167} &
        \multicolumn{3}{l}{Mean: 6.28 instances/image} \\
        \bottomrule
    \end{tabular}
        \caption{Distribution of WADE's 13,608 annotated instances.}
    \label{tab:wade_stats}
\end{table}

\subsection{Annotations and Validation}
\label{sec:annotations}

Four members of the author team exhaustively annotated every
identifiable target object using a dedicated annotation interface.
Each annotation contains an object-level bounding box and one of the
ten taxonomy labels. Two additional author-team members independently
reviewed the boxes and labels, correcting inaccurate or inconsistent
annotations. Overall, 95\% of the annotations were accepted without
modification.

Rather than authoring a separate explanation for every bounding box,
WADE defines one scene-invariant reasoning chain for each category.
Each chain represents category-level recognition knowledge, including
the category's expected appearance, primary visual cues, likely
confusions, contrastive decision rules, fallback evidence under partial
visibility, and common failure modes. The chain is associated with
every instance of its category, while the source image, bounding box,
and category label provide instance-specific visual and spatial
information.

This design avoids thousands of largely repetitive explanations and
provides consistent discriminative supervision for prompting and
fine-tuning. However, the chains should be interpreted as category-level
recognition guidance rather than fully image-grounded explanations of
individual instances. The complete reasoning-chain schema, annotation
protocol, and validation procedure are provided in
Appendix~\ref{app:annotation_details}.

\begin{figure}[t]
    \centering
    \includegraphics[width=\columnwidth]{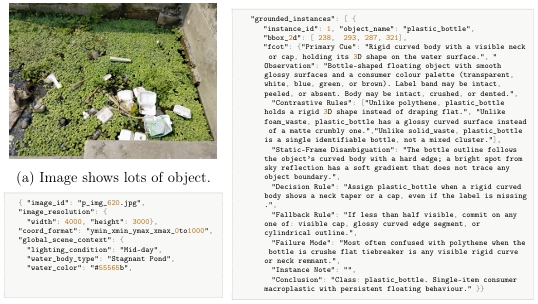}
    \caption{WADE annotation example. Every visible target instance receives
    a bounding box and category label and is linked to its class-level
    reasoning chain. Selected reasoning fields are shown for readability.}
    \label{fig:annotation_example}
\end{figure}

\subsection{Dataset Splits}
\label{sec:splits}

We construct training, validation, and test partitions using a
75/10/15 ratio, producing 1,625, 217, and 325 images, respectively.
Partitioning is performed at the water-body level: all images and
annotations collected from the same pond, canal, or river site remain
within a single partition. Consequently, no collection site appears
across the training, validation, and test sets, making the evaluation
both image-disjoint and location-disjoint at the collection-site level.
Additional collection and annotation details are provided in
Appendix~\ref{app:dataset_details}.

\section{Experimental Setup}
\label{sec:experimental_setup}
\begin{figure*}[t]
    \centering
    \includegraphics[width=\textwidth]{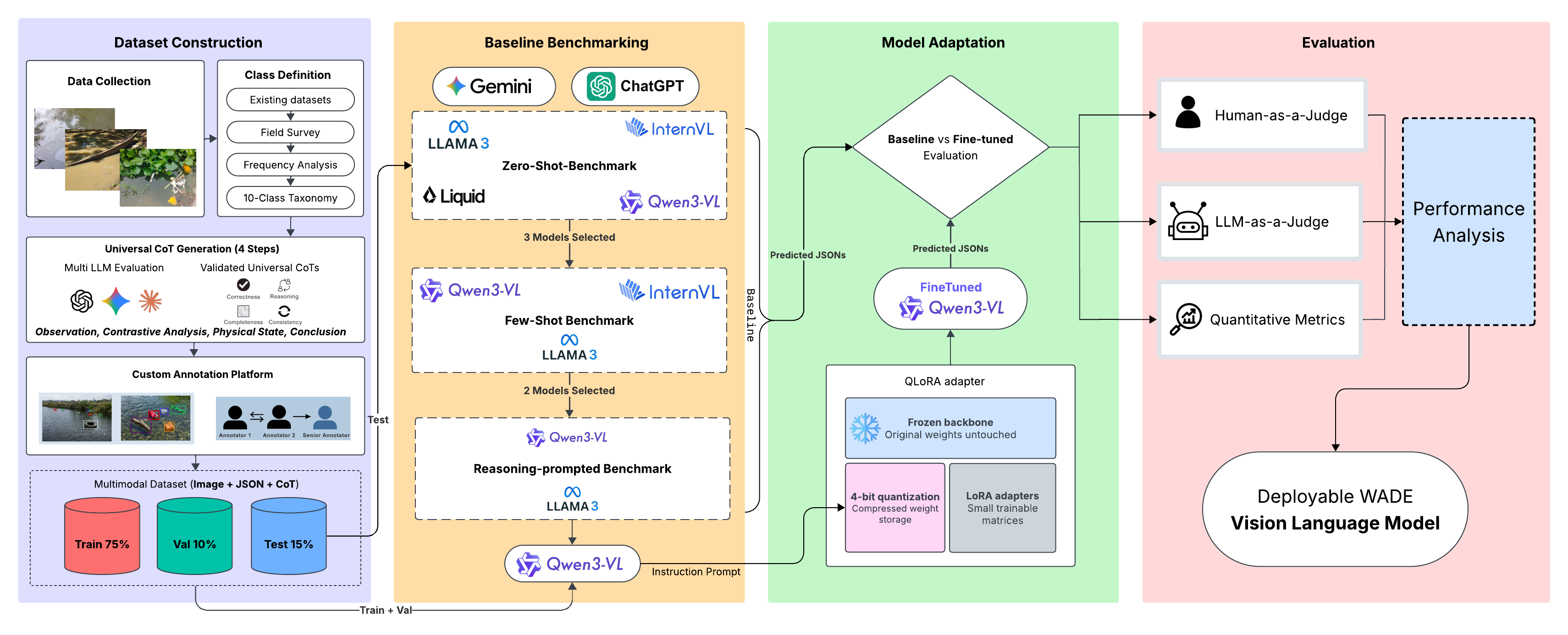}
    \caption{Overview of the WADE methodology. Field images are exhaustively
    annotated with bounding boxes, taxonomy labels, and class-level reasoning
    chains. Six vision-language models are evaluated under zero-shot,
    two-shot, reasoning-prompted, and QLoRA fine-tuning regimes. Predictions
    are assessed using automatic metrics, LLM judges, and blind human
    evaluation.}
    \label{fig:methodology}
\end{figure*}
We formulate WADE as structured generative grounding. Given an image, a
vision-language model must enumerate the visible target instances and return,
for each instance, a normalized bounding box, taxonomy label, and associated
class-level reasoning chain. We evaluate six VLMs under four regimes:
zero-shot inference, two-shot prompting, reasoning-guided prompting, and
reasoning-supervised QLoRA fine-tuning.

\subsection{Models and Output Format}
\label{sec:models}

Our evaluation includes four open-weight models:
LLaMA-3.2-11B-Vision, InternVL2-2B, Qwen3-VL-2B-Instruct, and
LFM2.5-VL-1.6B, together with two commercial API-based models, GPT-4o-mini and Gemini-2.5-Flash. The model set covers compact architectures between 1.6B and 2B parameters, a larger 11B model, and proprietary systems with undisclosed parameter counts. The compact models are included to examine whether computationally accessible VLMs can perform dense multi-instance grounding in low-resolution waterway images, while LLaMA-3.2-11B-Vision provides a larger-scale open-weight comparison. GPT-4o-mini and Gemini-2.5-Flash serve as commercial reference systems for evaluating whether stronger general-purpose multimodal capabilities transfer to this specialized environmental domain. Overall, this selection supports comparisons across model scale, accessibility, and adaptation capability.
\\
All models receive the same task instruction and output schema. A prediction
contains a list of grounded instances, with boxes represented as
$[y_{\min},x_{\min},y_{\max},x_{\max}]$ and normalized to $[0,1000]$. Each
instance additionally contains one of the ten WADE labels and a structured
reasoning chain. Requiring JSON output permits automatic parsing and
evaluation across model families. Malformed responses are counted as parse
failures and receive zero true positives rather than being removed from the
evaluation.

\subsection{Evaluation Regimes}
\label{sec:regimes}

\textbf{Zero-shot:}
We adopt a strict instruction-only zero-shot evaluation protocol, related to established zero-shot and open-vocabulary object-localization settings \cite{bansal2018zero,minderer2022simple}. Each model receives a single test image and the task instruction, without training examples, in-context demonstrations, or additional category descriptions. This protocol evaluates the model’s pretrained ability to recognize and spatially localize densely distributed floating waste.
\\ 
\textbf{Two-shot:}
Following multimodal in-context learning, the prompt is augmented with two
fixed demonstrations from the training set \cite{alayrac2022flamingo}.
Each demonstration includes an image and its complete ground-truth output,
comprising bounding boxes, labels, and reasoning chains. Because multimodal
in-context performance can be sensitive to demonstration selection
\cite{luo2024textual}, the same two demonstrations are used for every test
image to control for exemplar-induced variation. This regime is evaluated
with LLaMA-3.2-11B-Vision, InternVL2-2B, and Qwen3-VL-2B-Instruct, for which
the complete multimodal prompt fits within the available context budget.
\\
\textbf{Knowledge-guided prompting:}
Motivated by prior findings that discriminative class descriptions and
fine-grained visual descriptors can improve zero-shot recognition and
region--text alignment \cite{pratt2023platypus,jin2024llms}, the prompt
provides class-level recognition rules for all ten categories, including
their primary visual cues, likely confusions, and contrastive decision rules.
No solved image is supplied, and the model parameters remain unchanged.
This setting therefore isolates whether textual domain knowledge alone
improves visual grounding. We apply it to LLaMA-3.2-11B-Vision and
Qwen3-VL-2B-Instruct.
\\
\textbf{Reasoning-supervised fine-tuning:}
We fine-tune Qwen3-VL-2B-Instruct using QLoRA~\cite{dettmers2023qlora}.
The target is the complete ground-truth JSON containing every instance's box,
label, and associated class-level chain. Instances are ordered by
$y_{\min}$ before serialization to provide a consistent target sequence.
Loss is applied only to the answer tokens, while the instruction and image
tokens are masked. The vision encoder and base language model remain frozen,
and only the low-rank adapters are optimized. We use the checkpoint with the
lowest validation loss. Full training settings are reported in Appendix~\ref{app:training_details}.

Because fine-tuning jointly uses bounding boxes, category labels, and
reasoning chains, the observed gains reflect the combined effect of
in-domain WADE supervision. Without a box-and-label-only ablation, the
independent contribution of reasoning cannot be quantified.
\begin{table*}[t]
    \centering
    \scriptsize
    \setlength{\tabcolsep}{3.2pt}
    \begin{tabular}{llcccccccccc}
        \toprule
        & & \multicolumn{6}{c}{\textbf{Grounding}} &
        \multicolumn{2}{c}{\textbf{Hallucination}} &
        \multicolumn{2}{c}{\textbf{Counting Error}} \\
        \cmidrule(lr){3-8}
        \cmidrule(lr){9-10}
        \cmidrule(lr){11-12}

        \textbf{Regime} & \textbf{Model} &
        $\mathbf{P_{.5}}\uparrow$ &
        $\mathbf{R_{.5}}\uparrow$ &
        $\mathbf{F1_{.5}}\uparrow$ &
        $\mathbf{F1_{\mathrm{ag}}}\uparrow$ &
        $\mathbf{R_{.75}}\uparrow$ &
        $\mathbf{N_p/N_g}$ &
        $\mathbf{CHAIR_i}\downarrow$ &
        $\mathbf{CHAIR_s}\downarrow$ &
        $\mathbf{MAE}\downarrow$ &
        $\mathbf{RMSE}\downarrow$ \\
        \midrule

        Zero-shot & LLaMA-3.2-11B$\dagger$
        & .0070 & .0066 & .0068 & .0123 & .0027 & .94
        & .7621 & .9839 & 2.187 & 3.610 \\

        Zero-shot & InternVL2-2B$\dagger$
        & .0022 & .0046 & .0030 & .0043 & .0000 & 2.09
        & .8639 & .9960 & 4.325 & 5.232 \\

        Zero-shot & Qwen3-VL-2B$\dagger$
        & .0267 & .0248 & .0257 & .0347 & .0107 & .93
        & .6722 & .6836 & 2.540 & 3.889 \\

        Zero-shot & LFM2.5-VL-1.6B$\dagger$
        & .0031 & .0048 & .0043 & .0110 & .0019 & 1.55
        & .8374 & .5291 & 3.430 & 5.739 \\

        Zero-shot & GPT-4o-mini$\ddagger$
        & .0339 & .0153 & .0211 & .0293 & .0013 & .45
        & .5817 & .4679 & 1.954 & 3.597 \\

        Zero-shot & Gemini-2.5-Flash$\ddagger$
        & .1225 & .1127 & .1174 & .2008 & .0545 & .92
        & .6561 & .7822 & \textbf{1.366} & \textbf{1.893} \\

        \midrule

        Two-shot & LLaMA-3.2-11B$\dagger$
        & .0100 & .0122 & .0110 & .0150 & .0010 & 1.22
        & .6721 & .3876 & 2.725 & 4.426 \\

        Two-shot & InternVL2-2B$\dagger$
        & .0201 & .0030 & .0052 & .0087 & .0010 & .15
        & .7651 & .3855 & 3.424 & 5.525 \\

        Two-shot & Qwen3-VL-2B$\dagger$
        & .0085 & .0070 & .0077 & .0096 & .0052 & .82
        & .7176 & .5278 & 3.069 & 4.111 \\

        \midrule

        Reasoning & LLaMA-3.2-11B$\dagger$
        & .0149 & .0108 & .0125 & .0136 & .0072 & .72
        & .6791 & .9381 & 2.218 & 4.155 \\

        Reasoning & Qwen3-VL-2B$\dagger$
        & .0385 & .0137 & .0202 & .0403 & .0082 & .36
        & .6692 & .3077 & 2.523 & 3.610 \\

        \midrule

        Fine-tuned & Qwen3-VL-2B$\dagger$
        & \textbf{.2011} & \textbf{.2339} &
          \textbf{.2163} & \textbf{.2692} &
          \textbf{.1041} & 1.16 &
          \textbf{.2402} & \textbf{.0883} &
          2.934 & 4.214 \\

        \bottomrule
    \end{tabular}

    \caption{Grounding, hallucination, and counting performance on the WADE
    test set. Grounding metrics use class-aware matching unless marked
    class-agnostic ($F1_{\mathrm{ag}}$). $N_p/N_g$ is the ratio between
    predicted and ground-truth instance counts. CHAIR measures unsupported
    predictions. $\dagger$ denotes open-source and $\ddagger$ commercial
    models. Best values are shown in bold.}
    \label{tab:main_results}
\end{table*}

\subsection{Evaluation Protocol}
\label{sec:evaluation_protocol}
Predicted and ground-truth boxes are matched one-to-one by descending
intersection-over-union (IoU). We report class-aware precision, recall, and
F1 at $\operatorname{IoU}\geq0.5$, together with class-agnostic F1 and recall
at $\operatorname{IoU}\geq0.75$. Mean IoU is computed only over matched
pairs and therefore describes localization quality conditional on a match.

Because the generative models output textual boxes without confidence scores,
we do not report average precision, which requires ranking predictions by
confidence. Counting performance is measured using mean absolute error (MAE)
and root mean squared error (RMSE). Hallucination is measured using
instance-level and image-level CHAIR~\cite{rohrbach2018chair}.

Three LLM judges compare predicted and ground-truth JSON, while
two external, model-blind human evaluators visually compare
anonymized predictions against the source images without access
to ground-truth annotations. The external evaluators were not
involved in dataset construction, annotation validation, or model
development. Full protocols are provided in
Appendix~\ref{app:judge_protocol}.

\section{Results and Analysis}
\label{sec:results}
We organize the results around three questions: whether current VLMs can
ground dense floating waste without adaptation, whether additional prompting
improves their performance, and how strongly domain-specific fine-tuning
changes their behaviour.

\subsection{Dense Grounding Performance}
\label{sec:grounding_results}
Table~\ref{tab:main_results} reports grounding performance on the WADE
test set. All zero-shot models perform poorly at $\operatorname{IoU}=0.5$,
demonstrating that dense floating-waste grounding is difficult even for large
and commercial VLMs. Gemini-2.5-Flash is the strongest zero-shot model, with
an F1 of 0.1174, while Qwen3-VL-2B is the strongest compact open model, with
an F1 of 0.0257.

The low recall does not always result from generating too few predictions.
Qwen3-VL-2B, LLaMA-3.2-11B, and Gemini-2.5-Flash produce box-count ratios
of 0.93, 0.94, and 0.92, respectively. Thus, they generate approximately as
many boxes as the number of annotated objects, but only a small fraction are
correctly localized. For example, Qwen3-VL-2B produces 93\% of the
ground-truth count but recalls only 2.48\% of instances. Removing the class
constraint raises its F1 only from 0.0257 to 0.0347, indicating that incorrect
localization, rather than label disagreement alone, is the dominant failure.

The models also exhibit different generation behaviours. GPT-4o-mini is
conservative, producing only 0.45 boxes per annotated instance and obtaining
low recall. Conversely, InternVL2-2B produces more than twice the annotated
count but achieves only 0.0022 precision. Performance decreases further at
$\operatorname{IoU}=0.75$, showing that even some matched boxes are only
loosely aligned.

\subsection{Prompting Encourages Conservative Outputs}
\label{sec:prompting_results}

Two-shot prompting does not consistently improve grounding. Qwen3-VL-2B
declines from 0.0257 zero-shot F1 to 0.0077, while InternVL2-2B's box-count
ratio collapses from 2.09 to 0.15. LLaMA-3.2-11B is the only model whose
recall increases, from 0.0066 to 0.0122, but its precision remains close to
zero. These results suggest that examples of the required output structure are
insufficient to recover dense visual localization.

Reasoning-guided prompting produces a clearer precision--recall trade-off.
For Qwen3-VL-2B, precision improves from 0.0267 to 0.0385 and
class-agnostic F1 increases from 0.0347 to 0.0403. Its image-level
hallucination rate also falls from 0.6836 to 0.3077. However, recall decreases
from 0.0248 to 0.0137, while the box-count ratio falls from 0.93 to 0.36.
LLaMA-3.2-11B shows a similar reduction in output count. Supplying
discriminative rules therefore makes predictions more conservative but does
not recover the many small or occluded instances missed by the models.

\subsection{Effect of Reasoning-Supervised Fine-Tuning}
\label{sec:finetuning_results}

Fine-tuning produces the strongest improvement across the evaluated regimes.
Relative to zero-shot Qwen3-VL-2B, precision increases from 0.0267 to 0.2011,
while recall increases from 0.0248 to 0.2339, corresponding to approximately
$7.5\times$ and $9.4\times$ gains. F1 increases from 0.0257 to 0.2163, and
recall at the stricter IoU threshold of 0.75 rises from 0.0107 to 0.1041.
The predicted-to-ground-truth count ratio becomes 1.16, indicating slight
over-generation but substantially better spatial grounding.

The fine-tuned 2B model also exceeds Gemini-2.5-Flash, the strongest
zero-shot commercial system, in precision (0.2011 versus 0.1225), recall
(0.2339 versus 0.1127), and F1 (0.2163 versus 0.1174). This result
demonstrates the value of domain-specific supervision for compact VLMs.
However, because our current experiments do not include box-and-label-only
fine-tuning, the improvement must be attributed to the complete WADE
supervision package rather than to the reasoning chains alone.

\subsection{Hallucination and Counting}
\label{sec:main_results}

Table~\ref{tab:main_results} reports hallucination and counting
errors. Zero-shot models frequently generate unsupported objects.
InternVL2-2B and LLaMA-3.2-11B produce at least one hallucinated instance
in 99.60\% and 98.39\% of test images, respectively. Zero-shot
Qwen3-VL-2B obtains $\mathrm{CHAIR}_s=0.6836$.

Fine-tuning reduces Qwen3-VL-2B's instance-level CHAIR from 0.6722 to
0.2402 and image-level CHAIR from 0.6836 to 0.0883. Therefore, fewer than
9\% of test images contain an unsupported prediction after fine-tuning,
compared with more than 68\% before adaptation.

This improvement does not extend to counting accuracy. Fine-tuned
Qwen3-VL-2B obtains an MAE of 2.934, compared with 2.540 before
fine-tuning and 1.366 for Gemini-2.5-Flash. Correct enumeration and correct
localization are therefore partially separable: a model may predict a
reasonable number of boxes while placing them incorrectly, or improve
grounding without achieving the lowest count error.

\subsection{Judge and Human Evaluation}
\label{sec:judged_results}

The judge-based and human assessments support the automatic results. The
fine-tuned model ranks first on box quality, reasoning quality, and
hallucination control under both protocols. The LLM panel assigns it scores
of 6.8, 8.5, and 7.4, respectively, while the two blind human raters assign
7.5, 7.5, and 8.5.

Across all systems, reasoning scores are higher than box-quality scores. For
example, Gemini-2.5-Flash receives 7.8 for reasoning but only 4.1 for box
quality from the LLM panel. Similarly, reasoning-prompted Qwen3-VL-2B
receives 8.0 for reasoning but 4.3 for boxes. This confirms the central
quantitative finding: current VLMs can produce plausible descriptions of
waste categories more reliably than they can spatially ground the relevant
objects.

The judge and human evaluations agree on the overall model ranking but differ
in score magnitude. We therefore treat judge scores as comparative evidence
rather than absolute measurements. The complete judge and human results are
reported in Appendix~\ref{app:judge_results}.

\subsection{Remaining Difficulty}
\label{sec:remaining_difficulty}

Despite the improvement from fine-tuning, WADE remains far from solved. The
best model recalls only 23.39\% of instances at IoU 0.5 and 10.41\% at IoU
0.75. In other words, it misses more than three-quarters of annotated objects
under the standard threshold and nearly nine out of ten under stricter
localization. The remaining gap is particularly important for environmental
monitoring, where missed small or partially occluded waste objects may
accumulate across a waterway. These results position WADE as a challenging
benchmark for future work on dense, low-resolution grounding rather than a
saturated detection task.

\section{Discussion}
\label{sec:discussion}

Our experiments reveal a substantial gap between describing an object and spatially grounding it. Several zero-shot models generate approximately the correct number of boxes, yet only a small fraction overlap the annotated instances. Relaxing class-aware matching produces only limited improvement, indicating that the central difficulty is inaccurate localization rather than
taxonomy confusion alone. This behaviour is likely associated with the visual characteristics of WADE: objects are frequently small, crowded, partially occluded, and weakly separated from water, vegetation, and reflections.

Prompting does not resolve this limitation. Two-shot examples generally decrease performance, suggesting that demonstrations of the output format cannot compensate for inadequate visual grounding. Reasoning-guided prompting
improves precision and hallucination control for Qwen3-VL-2B but simultaneously reduces recall and the number of generated boxes. Providing discriminative rules therefore makes the model more selective: it produces fewer, more
cautious predictions while missing additional objects. In dense environmental scenes, this precision--recall trade-off is undesirable because exhaustive enumeration is essential for reliable monitoring.

Domain-specific fine-tuning produces a different outcome. Fine-tuned Qwen3-VL-2B improves precision, recall, strict-IoU recall, and hallucination control simultaneously, outperforming the strongest evaluated commercial
model despite its compact size. This result suggests that, for specialized environmental perception, targeted supervision can be more valuable than model scale alone. Nevertheless, the current experiment jointly supervises bounding boxes, labels, and class-level reasoning chains. Because we do not include a box-and-label-only fine-tuning condition, the improvement should be attributed to the complete WADE supervision package rather than to reasoning
annotations in isolation.

The judged evaluations expose a second important gap. Both LLM judges and human raters assign higher scores to generated reasoning than to bounding-box quality. A model can articulate a plausible visual cue or contrastive rule
even when its predicted coordinates are inaccurate. Fluent reasoning should therefore not be treated as evidence of successful visual grounding; geometric metrics and qualitative inspection remain necessary.

Finally, WADE is far from saturated. The best system recalls only 23.39\% of instances at IoU 0.5 and 10.41\% at IoU 0.75. These results establish a useful baseline while leaving substantial room for improvements in small-object
representation, resolution-aware visual encoding, dense-instance enumeration, and localization-aware multimodal training. Future experiments with
box-and-label-only supervision, dedicated detectors, and geographically disjoint evaluation will be necessary to determine which components drive generalization and whether the observed gains transfer to unseen waterways.

\section{Limitations}
\label{sec:limitations}

WADE has several limitations that define the scope of its current findings. First, the fine-tuning experiment jointly supervises bounding boxes, category labels, and class-level reasoning chains. Because a box-and-label-only fine-tuning condition was not evaluated, the contribution of reasoning
annotations cannot be isolated from the broader effect of domain-specific supervision. The current study also does not include comparisons with specialized detectors such as YOLO or DETR; consequently, the results characterize generative VLM grounding rather than superiority over task-specific detection architectures.

Second, the dataset split is location-disjoint at the collection-site level, such that images from the same pond, canal, or river site cannot occur across different partitions. However, all collection sites are located
in rural Bangladesh and may share broader regional characteristics. Consequently, the current evaluation does not establish generalization to other countries, urban drainage systems, coastal environments, or different image-capture platforms.

Third, WADE uses one scene-invariant reasoning chain per category. This provides consistent and scalable supervision but does not capture instance-specific factors such as unusual deformation, severe occlusion, or scene-dependent ambiguity. Although boxes and labels were independently
reviewed and 95\% were accepted without modification, we do not report a formal inter-annotator agreement coefficient.

Finally, the judge panel overlaps with the models used to validate the reasoning annotations, and the two human raters are members of the project team. These conditions may introduce evaluator bias, so judged scores are treated as supporting comparative evidence rather than absolute measures. Results from commercial systems may also change as their underlying APIs are updated. Future work should include independent evaluators, detector baselines, controlled annotation ablations, and geographically disjoint
evaluation.

\section{Conclusion}
\label{sec:conclusion}

We introduced WADE, a benchmark for dense floating-waste grounding with compact vision-language models. WADE contains 2,167 field images and 13,608 annotated instances across ten categories, with each instance assigned a bounding box, category label, and associated class-level reasoning chain. The dataset captures naturally occurring, crowded waterway scenes containing small objects, occlusion, reflections, vegetation, and substantial visual ambiguity. Our evaluation of six VLMs shows that existing models struggle primarily with spatial grounding. Zero-shot models often generate a plausible number of objects but place their boxes incorrectly, while two-shot and reasoning-guided prompting do not consistently improve recall. Reasoning-supervised QLoRA fine-tuning substantially improves precision, recall, strict-IoU localization, and hallucination control, allowing a compact 2B model to outperform the strongest evaluated commercial zero-shot system. However, the best model still recalls fewer than one-quarter of the annotated instances at IoU 0.5, demonstrating that WADE remains far from solved.

{
    \small
    \bibliographystyle{ieeenat_fullname}
    \bibliography{main}
}


\appendix

\section{Extended WADE Dataset Details}
\label{app:dataset_details}

\subsection{Collection Conditions}

WADE contains 2,167 images collected from approximately 500 ponds, canals,
and rivers across 390 geographic locations in rural Bangladesh. Most original
images were captured at a resolution of $3096\times4128$ pixels using consumer
smartphones. The use of readily available cameras reflects a practical
monitoring setting rather than a specialized sensor configuration.

The collection covers two major seasonal conditions: 1,000 images were
captured during winter and 1,167 during the monsoon season. It also includes
680 morning, 950 midday, and 537 low-light images. These variations introduce
reflections, shadows, turbidity, illumination changes, vegetation overlap,
partial occlusion, and weak object boundaries. No waste was introduced,
repositioned, or staged for photography.

In addition to object annotations, the annotation interface records four
image-level attributes: lighting condition, water-body type, primary
pollutant, and remediation priority. These attributes provide contextual
information about the captured environment but are not used as object
categories.

\subsection{Taxonomy Design}

The taxonomy was constructed around commonly observed surface materials and
their visually confusable alternatives. Water hyacinth and algal bloom are
retained even though they are biological materials because they commonly
coexist with anthropogenic waste and may obscure or resemble waste instances.
Similarly, industrial waste is separated from general solid waste to represent
differences in visual composition and potential remediation requirements.

The class distribution is naturally imbalanced. Water hyacinth and plastic
bottles are the most frequent classes, whereas foam waste and wood debris
appear less frequently. We preserve this distribution to reflect the observed
environment rather than balancing the dataset through removal or synthetic
resampling.

\section{Annotation and Reasoning Details}
\label{app:annotation_details}

\subsection{Bounding-Box Protocol}

Annotators were instructed to box every identifiable target instance,
including small, partially occluded, and overlapping objects. Bounding boxes
were drawn tightly around the visible extent of each object. Objects were not
annotated when their visible evidence was insufficient to assign a reliable
class. Each record stores the image identifier, image resolution, normalized
bounding-box coordinates, category label, and associated class-level reasoning
chain.

Two additional authors independently reviewed the completed annotations using
a separate validation interface. The review covered box placement, category
assignment, and image-level attributes. Incorrect boxes were adjusted,
questionable labels were reconsidered, and duplicate or inconsistent
annotations were corrected. In total, 95\% of annotations were accepted
without modification.

\subsection{Reasoning-Chain Schema}

WADE defines one validated reasoning chain for each of its ten
categories. As described in Section~\ref{sec:annotations}, these chains
encode class-level discriminative knowledge and are shared across
instances of the corresponding category. This section provides the
complete reasoning-chain structure.
Each chain follows a nine-field structure:

\begin{enumerate}
    \item \textbf{Primary Cue}: the most discriminative visual property;
    \item \textbf{Observation}: the category's expected appearance;
    \item \textbf{Contrastive Rules}: differences from confusable classes;
    \item \textbf{Static-Frame Disambiguation}: separation from reflections
          or other imaging artefacts;
    \item \textbf{Decision Rule}: the principal classification condition;
    \item \textbf{Fallback Rule}: evidence used under partial visibility;
    \item \textbf{Failure Mode}: the most likely source of confusion;
    \item \textbf{Instance Note}: optional instance-specific information; and
    \item \textbf{Conclusion}: a concise category summary.
\end{enumerate}

For example, the reasoning chain for \textit{plastic bottle} identifies a
rigid curved body, neck, or cap as its primary cue. Its contrastive rules
separate bottles from flexible polythene, matte or fragmented foam, and mixed
solid-waste clusters. The fallback rule permits classification from a visible
cap, rigid curved edge, or cylindrical outline when the object is partially
occluded.

\subsection{Reasoning Validation}

Because textual reasoning cannot be validated using bounding-box agreement,
each class-level chain was independently assessed by three language models.
Every judge assigned a score between 0 and 10. A chain was accepted only when
all three scores were at least 8 and their mean was at least 9. Chains failing
either condition were revised and resubmitted until they passed the gate.

This procedure checks whether the cues are visually meaningful, whether the
contrastive rules distinguish likely confusions, and whether the chain is
internally consistent. However, this validation should not be interpreted as
a replacement for human evaluation; it provides a scalable quality-control
mechanism for the current class-level rationales.

\subsection{Split Construction}

Images are grouped according to their water-body collection site before
partitioning. The site groups are assigned to training, validation, and
test sets using an approximately 75/10/15 ratio, resulting in 1,625
training images, 217 validation images, and 325 test images. All images
and annotations from the same pond, canal, or river site remain within
one partition. Therefore, the benchmark prevents image-level and
collection-site-level overlap across the three partitions.

\section{Extended Experimental Details}
\label{app:experimental_details}

\subsection{Prompt and Serialization}
\label{app:prompt_details}

All evaluation regimes use the same base instruction, coordinate convention,
taxonomy, and JSON schema. The conditioning information changes according to
the regime, but the required prediction format remains fixed. For every
visible target object, the model must generate an entry containing an instance
identifier, category label, normalized bounding box, and structured reasoning
chain.

Bounding boxes follow the format

\[
b_i =
[y_{\min},x_{\min},y_{\max},x_{\max}],
\qquad b_i \in [0,1000]^4.
\]

Normalizing the coordinates allows predictions from models using different
internal image resolutions to be evaluated under the same coordinate system.
Before fine-tuning, ground-truth instances are sorted from top to bottom using
$y_{\min}$. This ordering produces a consistent autoregressive target without
changing the underlying set of annotations.

A response is considered parseable when it can be converted into the required
JSON structure and its predicted instances contain valid coordinates and
category fields. Responses that cannot be parsed are retained as failures
with zero true positives. This prevents malformed generations from being
silently excluded.

\subsection{Evaluation-Regime Details}
\label{app:regime_details}

\paragraph{Zero-shot inference.}
The input consists of the task instruction and one test image. No
demonstrations, category-specific reasoning rules, or parameter updates are
provided. We evaluate LLaMA-3.2-11B-Vision, InternVL2-2B,
Qwen3-VL-2B-Instruct, LFM2.5-VL-1.6B, GPT-4o-mini, and
Gemini-2.5-Flash in this setting.

\paragraph{Two-shot prompting.}
Two solved training examples are placed before the test query. Each example
contains an image and its complete serialized annotation. The exemplars remain
fixed across models and test images. We restrict this experiment to
LLaMA-3.2-11B-Vision, InternVL2-2B, and Qwen3-VL-2B-Instruct because the
combined images and annotations exceed the practical context budget of the
remaining evaluated systems.

\paragraph{Reasoning-guided prompting.}
For every WADE category, the prompt includes the primary cue, confusable
category, and contrastive recognition rule. No example image or solved
annotation is included. Thus, the regime evaluates the effect of supplying
textual domain knowledge during inference without updating model parameters.
We evaluate Qwen3-VL-2B-Instruct and LLaMA-3.2-11B-Vision in this setting.

\subsection{QLoRA Configuration}
\label{app:training_details}

Qwen3-VL-2B-Instruct is loaded using 4-bit NF4 quantization with double
quantization and bfloat16 computation. Low-rank adapters use rank $r=16$,
scaling parameter $\alpha=32$, and dropout $0.05$. The adapters are attached
to the language model's attention projections
(\texttt{q}, \texttt{k}, \texttt{v}, and \texttt{o}) and feed-forward
projections (\texttt{gate}, \texttt{up}, and \texttt{down}). The vision
encoder and remaining backbone parameters stay frozen.

The training target is a single serialized JSON sequence containing all
ground-truth objects in an image. The loss is masked over the prompt and image
tokens and calculated only over the target response. Training uses the
following configuration:

\begin{table}[t]
    \centering
    \caption{QLoRA fine-tuning configuration.}
    \label{tab:training_configuration}
    \small
    \begin{tabular}{lc}
        \toprule
        \textbf{Hyperparameter} & \textbf{Value} \\
        \midrule
        Base model              & Qwen3-VL-2B-Instruct \\
        Quantization            & 4-bit NF4 \\
        LoRA rank               & 16 \\
        LoRA $\alpha$           & 32 \\
        LoRA dropout            & 0.05 \\
        Epochs                  & 3 \\
        Learning rate           & $2\times10^{-4}$ \\
        Per-device batch size   & 1 \\
        Gradient accumulation   & 8 \\
        Effective batch size    & 8 \\
        Warmup ratio            & 0.03 \\
        Maximum sequence length & 4,096 \\
        Maximum vision tokens   & 1,024 \\
        Decoding                & Greedy \\
        Checkpoint selection    & Lowest validation loss \\
        \bottomrule
    \end{tabular}
\end{table}

Gradient checkpointing is enabled to reduce memory consumption. Only adapter
parameters are updated. We refer to this setting as
\emph{reasoning-supervised fine-tuning}, not knowledge distillation, because
the targets are derived from human-authored annotations rather than outputs
from a teacher model.
\subsection{Box Matching and Detection Metrics}
\label{app:metric_details}

For every image, pairwise intersection-over-union (IoU) is calculated between
each predicted box $b_p$ and ground-truth box $b_g$:

\[
\operatorname{IoU}(b_p,b_g)
=
\frac{|b_p\cap b_g|}
{|b_p\cup b_g|}.
\]

Candidate pairs are sorted by decreasing IoU and matched greedily under a
one-to-one constraint. In class-aware evaluation, a pair is eligible only
when its predicted and ground-truth labels agree. In class-agnostic
evaluation, category labels are ignored.

At threshold $\tau$, precision, recall, and F1 are defined as

\[
P_{\tau}=\frac{TP_{\tau}}{TP_{\tau}+FP_{\tau}},
\qquad
R_{\tau}=\frac{TP_{\tau}}{TP_{\tau}+FN_{\tau}},
\]

\[
F1_{\tau}
=
\frac{2P_{\tau}R_{\tau}}
{P_{\tau}+R_{\tau}}.
\]

We report class-aware $P_{0.5}$, $R_{0.5}$, and $F1_{0.5}$,
class-agnostic $F1_{0.5}$, and class-aware $R_{0.75}$. We do not report
average precision because the generative VLMs produce textual bounding boxes
without confidence scores, preventing confidence-based ranking and threshold
sweeping.

We additionally report the predicted-to-ground-truth count ratio,

\[
\frac{N_p}{N_g}
=
\frac{\sum_{i=1}^{N}\hat{n}_i}
     {\sum_{i=1}^{N}n_i},
\]

where $\hat{n}_i$ and $n_i$ denote the predicted and ground-truth instance
counts for image $i$, respectively. A value below one indicates
under-generation, whereas a value above one indicates over-generation.

Counting accuracy is measured using mean absolute error (MAE) and root mean
squared error (RMSE):

\[
\operatorname{MAE}
=
\frac{1}{N}\sum_{i=1}^{N}|\hat{n}_i-n_i|,
\]

\[
\operatorname{RMSE}
=
\sqrt{\frac{1}{N}\sum_{i=1}^{N}(\hat{n}_i-n_i)^2}.
\]

Hallucination is evaluated using CHAIR~\cite{rohrbach2018chair}. We report
the fraction of generated instances unsupported by the ground-truth
annotations, denoted by $\mathrm{CHAIR}_{i}$, and the fraction of images
containing at least one unsupported prediction, denoted by
$\mathrm{CHAIR}_{s}$. Lower values indicate better hallucination control.

\subsection{Judge-Based and Human Evaluation}
\label{app:judge_protocol}

\paragraph{LLM-as-judge evaluation.}
We use Gemini-3.5-Flash, GPT-5.4, and Claude-Opus-4.8 as
three separate judges. For each test image, every judge receives
the predicted JSON and corresponding ground-truth JSON. Model
identity and evaluation regime are removed, and the raw image is
not provided.

Each prediction is scored from 0 to 10 along three dimensions:
\begin{itemize}
    \item \textbf{Box agreement}: correspondence between the
    predicted and ground-truth instances and alignment of their
    serialized bounding-box coordinates;

    \item \textbf{Reasoning consistency}: agreement of the
    predicted cues, likely confusions, and discriminative rules
    with the reference class-level reasoning chains; and

    \item \textbf{Hallucination control}: the extent to which
    predicted instances correspond to ground-truth objects rather
    than unsupported predictions.
\end{itemize}

Scores are averaged across the test images and then across the
three judges. No acceptance threshold is applied. Because the
judges receive serialized annotations without the raw image, their
scores measure reference-based agreement rather than direct visual
correctness. We therefore treat them as comparative evidence
alongside the geometric grounding metrics.

\paragraph{Human comparative evaluation.}
Two external evaluators independently perform a comparative visual
evaluation of seven selected model--regime conditions: zero-shot
outputs from five models, together with the reasoning-guided and
fine-tuned Qwen3-VL-2B conditions. Few-shot outputs are not included.
The evaluators were not involved in dataset construction, annotation
validation, model development, or LLM-based judging.

For each test image, the evaluators are shown the source image alongside
the anonymized predictions, including bounding boxes, category labels,
and reasoning chains. Ground-truth annotations, model identities, and
evaluation-regime names are hidden. The evaluators compare the
predictions with one another and determine which outputs correspond
more accurately to the visible content of the source image.

After reviewing the complete test-set outputs, each evaluator assigns
a comparative score from 0 to 10 to every model--regime condition along
three dimensions:

\begin{itemize}
    \item \textbf{Box quality}: the relative accuracy and coverage of
    the predicted boxes over visible waste instances;

    \item \textbf{Reasoning quality}: the relative relevance and
    consistency of the generated reasoning with the visible objects
    and predicted categories; and

    \item \textbf{Hallucination control}: the relative ability to avoid
    predicting objects, labels, or boxes unsupported by the visible
    image.
\end{itemize}

Higher scores indicate better performance relative to the other
anonymized predictions. The evaluators rely only on the source images
and model predictions and do not access ground-truth annotations.
The final score for each model--regime condition is averaged across
the two evaluators. These scores therefore represent external
comparative visual assessments rather than absolute geometric accuracy.
\section{Judge and Human Evaluation Results}
\label{app:judge_results}

\begin{table}[h]
    \centering
    \scriptsize
    \setlength{\tabcolsep}{3pt}
    \begin{tabular}{llccc}
        \toprule
        \textbf{Regime} & \textbf{Model} &
        \textbf{Box} & \textbf{Reason.} & \textbf{Halluc.} \\
        \midrule
        Zero-shot & InternVL2-2B        & 2.1 & 4.0 & 2.5 \\
        Zero-shot & LLaMA-3.2-11B       & 3.0 & 5.5 & 2.6 \\
        Zero-shot & Qwen3-VL-2B         & 3.6 & 6.5 & 3.8 \\
        Zero-shot & GPT-4o-mini         & 3.9 & 7.0 & 5.1 \\
        Zero-shot & Gemini-2.5-Flash    & 4.1 & 7.8 & 4.2 \\
        Reasoning & Qwen3-VL-2B         & 4.3 & 8.0 & 5.8 \\
        Fine-tuned & Qwen3-VL-2B        & \textbf{6.8} &
          \textbf{8.5} & \textbf{7.4} \\
        \bottomrule
    \end{tabular}
        \caption{LLM-as-judge evaluation averaged over three judges.}
    \label{tab:llm_judge_results}
\end{table}

\begin{table}[h]
    \centering
    \scriptsize
    \setlength{\tabcolsep}{3pt}
    \begin{tabular}{llccc}
        \toprule
        \textbf{Regime} & \textbf{Model} &
        \textbf{Box} & \textbf{Reason.} & \textbf{Halluc.} \\
        \midrule
        Zero-shot & InternVL2-2B        & 1.5 & 3.5 & 2.5 \\
        Zero-shot & LLaMA-3.2-11B       & 3.0 & 4.0 & 3.5 \\
        Zero-shot & Qwen3-VL-2B         & 4.0 & 5.5 & 3.5 \\
        Zero-shot & GPT-4o-mini         & 4.5 & 6.0 & 6.5 \\
        Zero-shot & Gemini-2.5-Flash    & 5.0 & 7.0 & 6.5 \\
        Reasoning & Qwen3-VL-2B         & 4.5 & 6.0 & 4.5 \\
        Fine-tuned & Qwen3-VL-2B        & \textbf{7.5} &
          \textbf{7.5} & \textbf{8.5} \\
        \bottomrule
    \end{tabular}
        \caption{Blind human evaluation averaged over two independent raters.}
    \label{tab:human_results}
\end{table}
The fine-tuned model receives the strongest scores on every axis under both
evaluation protocols. The LLM panel assigns scores of 6.8, 8.5, and 7.4
for box quality, reasoning quality, and hallucination control, respectively.
Human raters assign corresponding scores of 7.5, 7.5, and 8.5.

Both protocols rank Gemini-2.5-Flash as the strongest zero-shot system and
the fine-tuned Qwen3-VL-2B as the strongest system overall. However, the
differences in score magnitude show that the judge panel is more favourable
toward fluent reasoning, while human raters are more sensitive to spatial
accuracy and unsupported detections. We therefore interpret judge scores
comparatively rather than as absolute quality measurements.

\end{document}